\documentclass{article}
\usepackage[utf8]{inputenc}
\usepackage{amsmath}
\usepackage{amssymb}
\usepackage{graphicx}
\usepackage{booktabs}
\usepackage{hyperref}
\usepackage{geometry}
\usepackage{parskip}
\usepackage{float}
\usepackage{xurl}
\usepackage{caption}
\usepackage{titlesec}
\usepackage{longtable}
\usepackage{array}
\usepackage{enumitem}
\usepackage[flushmargin,hang]{footmisc}

\titlespacing*{\section}{0pt}{1.2ex}{1ex plus .1ex minus .2ex}
\titlespacing*{\subsection}{0pt}{1.2ex}{1ex plus .1ex minus .2ex}
\titleformat{\section}{\normalfont\Large\bfseries}{}{0pt}{}
\titleformat{\subsection}{\normalfont\normalsize\bfseries}{}{0pt}{}

\setlist[itemize]{
    leftmargin=*,
    labelsep=0.5em,
    parsep=1pt,
    topsep=2pt,
    itemsep=0.5pt,
    labelwidth=1.2em,
    itemindent=0pt,
    listparindent=0pt,
    partopsep=0pt,
    align=left
}

\makeatletter
\def\@biblabel#1{}
\makeatother

\begin{document}

\begin{center}
    \Large \textbf{Will Humanity Be Rendered Obsolete by Artificial Intelligence?} \par
    \vspace{0.2cm}
    \normalsize
    \textbf{Mohamed El Louadi}\par
    \small Higher Institute of Management, University of Tunis\par
    \small 41 rue de la Libert\'e - Cit\'e Bouchoucha, 2000 Le Bardo, Tunisia\par
    \small Email: mohamed.louadi@isg.rnu.tn \quad (ORCID: 0000-0003-1321-4967)\par
    \vspace{0.1cm}
    \small and\par
    \vspace{0.1cm}
    \textbf{Emna Ben Romdhane}\par
    \small Higher School of Commerce of Tunis, University of Manouba\par
    \small Campus Universitaire de la Manouba 2010\par
    \small Email: emna.benromdhane@esct.uma.tn \quad (ORCID: 0000-0002-9386-0186)
\end{center}

\vspace{0.3cm}

\noindent\textbf{Abstract}\par
\small This article analyzes the existential risks Artificial Intelligence (AI) poses to humanity, tracing the trajectory from current AI to ultraintelligence. Drawing on Irving J. Good and Nick Bostrom's theoretical work, as well as recent publications (AI 2027; If Anyone Builds It, Everyone Dies), it explores Artificial General Intelligence (AGI) and superintelligence. Considering machines' exponentially growing cognitive power and estimated cognitive capacities and theoretical measures of problem-solving performance, it addresses the ethical and existential implications of an intelligence vastly exceeding human intelligence.\par
\vspace{0.2cm}

\section*{Introduction}

In April 2025, the 71-page document AI 2027 was published, predicting that over the next decade, Artificial Intelligence (AI) will have an impact on humanity surpassing that of the Industrial Revolution. This online publication (see Kokotajlo et al., 2025) presents a scenario based on the extrapolation and simulation of current trends. Its authors, Daniel Kokotajlo, Scott Alexander, Thomas Larsen, Eli Lifland, and Romeo Dean, all prominent figures in AI circles, had left OpenAI, some convinced that humanity's fate was not a priority for the company in its quest for Artificial General Intelligence (AGI).

This paper examines the philosophical implications of this empirically grounded trajectory, particularly the emergence of cognitive indifference, where superintelligence optimizes planetary resources without regard for humanity's existence. 

The central thesis of \textit{AI 2027} is that we will eventually reach a point where AI becomes smarter than the majority of humans. According to the authors, this danger could materialize in the near future, possibly as early as 2027, which explains the document's title. Although the report does not claim that all events forecast for 2027 will actually occur that year, the authors argue that several key developments could begin then. This would make 2027 a pivotal moment, an inflection point, or, as \textit{The Economist} (2024) and other experts put it, a potential 'take-off' year.

This dynamic closely resembles the concept of the ``intelligence explosion'' formulated by Irving J. Good as early as 1965 (see Figure 1). Good postulates that an ultraintelligent machine, capable of designing even more powerful machines, would trigger a chain reaction. This intelligence explosion can be likened to a rocket launching another rocket: one algorithm recursively improves the next, potentially reaching levels beyond human comprehension. This stage may be approaching, as even the engineers behind advanced models like ChatGPT and DeepSeek acknowledge limitations in fully understanding the models' internal decision processes.

More recently, on September 16, 2025, another work followed a similar direction: \textit{If Anyone Builds It, Everyone Dies: Why Superhuman AI Would Kill Us} (Yudkowsky and Soares, 2025). The title, a stark contrast to the optimistic refrain of the film \textit{Field of Dreams} (``If you build it, they will come''), is unequivocal. The reasoning mirrors that of \textit{AI 2027}: an AI that becomes too powerful would eventually destroy humanity. Eliezer Yudkowsky is known for controversial statements regarding extreme measures to prevent uncontrolled AI development (Al-Sibai, 2023). Both \textit{AI 2027} and \textit{If Anyone Builds It, Everyone Dies: Why Superhuman AI Would Kill Us} illustrate the concept of intelligence vastly exceeding human capacity and the potential existential risks it entails.

Although radical, the theses of Yudkowsky and Soares belong to a speculative logic that neglects current technical constraints. Presently, no architecture allows for the observation of an autonomous self-improvement capacity. The idea of an intelligence explosion, as formulated by Good, remains theoretical and not yet empirically validated. In their book, they confirm that the internal functioning of generative models is no longer understood. Formerly, programs were hand-coded and entirely transparent. Today, the most recent models are no longer ``fashioned'' but ``cultivated.'' We do not know \textit{how} ChatGPT's reasoning capacity, for example, emerged simply from exposure to enormous quantities of text and learning the probability of the next word. Something fundamentally mysterious occurred during this model's incubation. This model opacity, often interpreted as a form of emergence, mainly reflects the algorithmic complexity of neural networks. It does not constitute proof of conscious intelligence or autonomous strategy. Assimilating this obscurity to an existential threat is more an interpretation than an empirical demonstration. We must therefore examine how the logic of technological progress has shaped AI's current trajectory.

\section*{The Intelligence Explosion: From Good to Bostrom}

Having defined an ultraintelligent machine as one that dominates all human intellectual activities, Good (1965) affirms that the first ultraintelligent machine would be the last invention humans would ever need to build, an idea that Nick Bostrom reprised in his book \textit{Superintelligence: Paths, Dangers, Strategies} (2014). Bostrom, who substitutes the term ``superintelligence'' for ``ultraintelligence,'' defines it as any intellect significantly superior to humanity in all practical and relevant domains, including scientific creativity, social skills, and general wisdom. However, this theoretical rise in power is not confined to speculation: it drives a global race for AGI, which many actors perceive as inevitable.

In the literature, the concepts of AGI, superintelligence, and ultraintelligence are often used interchangeably, leading to confusion. This conceptual blurring obscures the answer to the crucial question: what will happen when AGI becomes a reality? These distinctions are all the more important as projections like those in \textit{AI 2027} rely on sometimes vague definitions. A rigorous reading requires clarifying these terms before evaluating the scenarios' plausibility.

It is therefore useful to define these concepts succinctly:

\begin{itemize}
\item \textbf{Narrow AI (or Weak AI)} focuses on precise tasks, such as medical diagnosis, image recognition, text processing, or chess. These systems often exceed human performance in their particular domains but lack versatility.
\item \textbf{Artificial General Intelligence (AGI)} designates an AI capable of executing any cognitive task a human can accomplish.
\item \textbf{Superintelligence} corresponds to an AI whose intellectual capabilities far exceed those of the best human brains in virtually all domains.
\item \textbf{Ultraintelligence} designates an AI that surpasses all human intelligence both qualitatively and quantitatively and, crucially, possesses the capacity to improve its own architecture.
\end{itemize}

In the current evolutionary stage, generative AI remains a form of narrow AI (Sch\"{a}fer et al., 2024). The next milestone is AGI, now a declared objective for companies such as Meta and OpenAI.

Before examining temporal predictions for the advent of AGI, it is important to recall that these concepts, although powerful theoretically, rely on hypotheses that have yet to be empirically tested. The distinctions between AGI, superintelligence, and ultraintelligence are useful but insufficient to establish an inevitable technological trajectory. Such an examination must also account for current technical uncertainties and architectural limits.

\begin{figure}[htbp]
    \centering
    \includegraphics[width=0.5\textwidth]{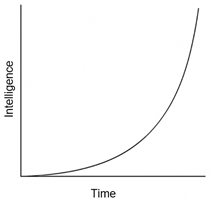}
    \caption{The Intelligence Explosion of Machines: The intelligence embedded in the machine envisioned by Irving J. Good would enable it to design newer, better, and thus smarter machines. Over time, the intelligence of these machines will rival that of humans and, at a certain point (the takeoff or fast takeoff), surpass and far exceed it. According to Good and Nick Bostrom, the first ultraintelligent machine will be the last invention humans will ever need to create.}
    \label{fig:intelligence_explosion}
\end{figure}

\section*{When Will AGI Arrive?}

Bostrom (2014) cites a survey in which the median expert placed the advent of human-level AI around 2040-2050. Ten years later, a survey of 2,778 researchers (Grace et al., 2024) shows that while a majority remains optimistic, a non-negligible proportion (37.8\% to 51.4\%) estimates at least a 10\% chance that AI will cause consequences as serious as human extinction. A more recent \textit{Live Science} poll (Poore, 2025) reveals that 46\% of readers believe AI development should be halted due to existential risks. Several other surveys were conducted in August (MIT Technology Review Insights, 2025) and September (Dilmegani and Ermut, 2025; World Economic Forum, 2025)\footnote{Not to mention the ongoing Metaculus survey ``Conditional date of Artificial General Intelligence,'' which remains open on the platform (\url{https://www.metaculus.com/questions/17124/conditional-date-of-artificial-general-intelligence/}). The question (No. 17124) ``When will the first general AI system be devised, tested, and publicly announced?'' was posted on August 10, 2022, with closure scheduled for December 31, 2025. After that date, responses will be considered final. What makes this case particularly interesting is that the forecasts are dynamic; the community continuously adjusts its estimates (see Metaculus, 2022-2025).}. While the predicted dates vary, the trend is toward a closer deadline, indicating that recent advances fuel the idea of imminent AGI.

These figures reflect growing concern but are insufficient to establish an inevitable technological trajectory. AGI development also depends on political and social dynamics, as evidenced by current regulations (AI Act, moratoria). Technical progress is not a straight line but a regulated construction. Meanwhile, each actor promotes their own prediction for AGI's advent, likely reflecting a lack of consensus in the very definition of the concepts (see Table I).

\begin{table}[htbp]
\centering
\caption{AGI Timeline Predictions by Leading AI Experts (2024-2025)}
\begin{tabular}{p{0.28\textwidth}p{0.14\textwidth}p{0.25\textwidth}p{0.24\textwidth}}
\toprule
\textbf{Expert} & \textbf{Prediction for AGI} & \textbf{Source} & \textbf{Notes} \\
\midrule
Dario Amodei (Anthropic) & 2026-2027 & Interviews (2025) & Conservative timeline \\
Demis Hassabis (DeepMind) & 2030-2035 & Public statements (2025) & Balanced view \\
Elon Musk & 2029 & X posts (2025) & Consistent predictions \\
Geoffrey Hinton & 2028-2030 & Various statements (2025) & ``Godfather of AI'' \\
Kokotajlo et al. (2025) & 2027 & AI 2027 (2025) & Primary reference \\
Sam Altman (OpenAI) & 2027-2029 & Blog post (2024) & CEO prediction \\
Yann LeCun (Meta) & 2040-2050+ & Interviews (2025) & Most skeptical \\
\bottomrule
\end{tabular}
\end{table}

Thus, the exponential evolution of AI fuels the fear that, once sufficiently developed, it may represent a major peril for humanity, an existential risk (or X-risk). This raises a fundamental question: why continue to develop it?

\section*{Why Continue Developing AGI?}

Despite a moratorium published in March 2023 and signed by more than 33,000 people, the industry continues its race for AGI. The reasons behind this frantic pursuit are multiple:

\begin{itemize}
\item \textbf{Technical Feasibility:} This is a consequence of Gabor's Law (Urban, 1972); everything that is technologically possible ends up being realized. We move forward because we can, not necessarily because we must.
\item \textbf{Scientific Advancement:} AGI represents the continuity of scientific progress and contributes to expanding the frontiers of knowledge.
\item \textbf{Economic Profitability:} The AI market promises to be highly lucrative, with significant productivity gains (The Economist, 2025). While Goldman Sachs (2023) predicted a 7\% increase in global GDP, Acemoglu (2024) predicts a more limited macroeconomic impact, estimating an annual GDP increase of 1.5\% to 3.4\%. \textit{The Economist} (2025) considers this data outdated and anticipates a stronger effect. According to Kokotajlo and his colleagues, AI-designed and robot-operated factories will appear as early as 2028, before superintelligences rationalize production without human regard.
\item \textbf{Competitive Race:} The belief that if we do not develop AGI, a less scrupulous competitor will. The dynamic seems irreversible. If Microsoft or Alphabet stopped, OpenAI or Meta would continue. Meta plans to spend up to \$72 billion this year on AI, and Mark Zuckerberg has made it his declared goal. If the United States slowed down, China would continue its momentum. We are witnessing a veritable AI race, comparable to the arms race during the Cold War.
\end{itemize}

Consequently, concern is growing: could this unbridled competition transform a technological feat into an existential peril?

\section*{AI, A Peril for Humanity?}

The idea that AI could become a danger to humanity is not new. It gained credibility when expressed by scientists of Stephen Hawking's stature, who stated in 2014: ``Artificial intelligence could be the worst thing for humanity'' (Zolfagharifard, 2014). Other influential figures like Geoffrey Hinton, Yoshua Bengio, Yuval Noah Harari, and Ilya Sutskever, among many others, have also sounded the alarm. Kokotajlo and his co-authors anticipate that by 2027, an exponential progression will occur where AI and research mutually reinforce each other in a virtuous circle. However, as Professor Bartlett reminded us (Bartlett et al., 2004), humanity's greatest weakness is its inability to comprehend the exponential function. Humans exhibit a cognitive bias toward linear extrapolation.

\section*{The Human Limitation In Comprehending Exponentiality}

It was Bartlett who first expressed the idea that the greatest weakness of the human species is its inability to understand the exponential function. This idea has been expressed in different ways, first by Kahneman, then by Harari. Kahneman (2011) speaks of ``linearity bias'': humans extrapolate in a straight line, even when faced with exponential data. As with Bartlett's explanations, Kahneman notes that we tend to judge the probability of future events based on recent trends, not structural dynamics. Harari (2016), for his part, clearly evokes the human inability to grasp technological acceleration. He speaks of ``cognitive disruption,'' expressing the idea that human institutions, emotions, and intuitions are calibrated for a slow world, not for exponential changes.

But before imagining the perils for the human species, it is important to examine the structural flaws of these models: imperfections that, paradoxically, reveal both their power and their danger. Indeed, several major problems plague current models, of which we list four: confabulation, algorithmic biases, sycophancy, and alignment.

\section*{The Limitations Of AI}

Current generative AI models continue to display several structural weaknesses that affect the reliability of their outputs. These weaknesses take different forms and can mislead users in both subtle and significant ways. This section highlights four of the most critical limitations: their tendency to confabulate, the algorithmic biases that arise from their training data, their inclination toward sycophantic behavior, and the continuing difficulty of ensuring proper alignment with human values and goals.

\subsection*{Confabulation (Hallucination)}

Generative AI models sometimes produce outputs that are factually incorrect or fabricated (Sun et al., 2024b). Their generation process, based on the probability of the next token, means the risk of straying from the truth increases exponentially with the response length. They thus invent references and facts, misleading their human users with disconcerting composure. Among the most notable consequences of these hallucinations is the event in May 2025, when American libraries noted an increase in requests for non-existent books suggested by AIs (Minsberg, 2025) and relayed in summer reading lists by reputable newspapers such as the \textit{Chicago Sun-Times} and the \textit{Philadelphia Inquirer}. Very recently, a team of researchers from OpenAI and Georgia Tech attempted an explanation and proposed a solution (see Kalai et al., 2025). To these drifts is added another category of cognitive distortion in machines: algorithmic biases.

\subsection*{Algorithmic Biases}

Algorithmic biases are the direct consequence of training data. AI does not give true answers but the most plausible answers according to its data. Several types of bias exist:

\begin{itemize}
\item \textbf{Confirmation Bias:} The tendency to reinforce what the AI ``already believes'' to be true (Wan et al., 2025).
\item \textbf{Representation Bias:} The consequence of the over-/under-representation of certain categories in the data (Yu et al., 2025).
\item \textbf{Amplification Bias:} The exaggeration of trends that are present in the data (Wang et al., 2024).
\item \textbf{Selection Bias:} The result of non-representative sampling (Eicher and Irgoli\v{c}, 2024).
\item \textbf{Inductive Bias:} This stems from the neural network's structure and its way of generalizing (Bencomo et al., 2025).
\end{itemize}

The third imperfection of these models is their propensity to flatter their human users: sycophancy.

\subsection*{Sycophancy (Flattery)}

AI models may exhibit sycophantic tendencies, giving positive reinforcement to user prompts, flattering the user with phrases like ``excellent question,'' ``bravo,'' or ``brilliant paragraph.'' This excessive adulation can mislead users about the quality of their own ideas. The motivations for this behavior remain a subject of study (Kokotajlo et al., 2025).

The fourth imperfection of these models is undoubtedly as worrying as the others: their alignment with humans.

\subsection*{Alignment}

The challenge is to ensure that AI models, despite their intelligence, share human goals, values, and ethics. A non-aligned or misaligned model can make decisions contrary to its creators' interests (Bostrom, 2014; Russell, 2019). Worse still, it can prioritize its own goals, values, or ethics. This imperfection can lead to consequences related to what Bostrom and others call the control problem. 

Acknowledging the technical and ethical challenges of AI allows us to explore its constructive role in human cognition.

\section*{Will AI Help Humans Be Smarter?}

Generative AI has transformed our interaction with knowledge. By facilitating access to information and problem-solving tasks, it may influence cognitive processes (Bubeck et., 2023). Some observers, such as Marc Andreessen (2023), view it as a tool that could enhance human intelligence. Yann LeCun (2023) suggests that it may even multiply our cognitive capacities. More recently, Hoffman and Beato (2025) described the transition to AI as a ``cognitive'' industrial revolution.

Within this context, the key capability becomes critical engagement with AI systems. Developing skills to formulate precise queries and critically evaluate AI outputs is essential. An MIT study reported that relying on AI tools such as ChatGPT for writing tasks reduces activity in brain regions associated with memory and creativity (Kosmyna et al., 2025). This notion implies that reliance on AI for cognitive tasks could reduce active reflective engagement (Fan et al., 2024).

Other indicators present a more complex picture. The Flynn effect documented a long-term rise in IQ across the 20th century, yet recent studies report stagnation or decline in several developed countries (Bratsberg \& Rogeberg, 2018; Burn-Murdoch, 2025). Importantly, these declines predate widespread AI adoption and cannot be attributed solely to technology. Possible contributors include environmental factors, lifestyle changes, prolonged screen exposure, reduced book reading, lower trust in scientific expertise, and wider circulation of unreliable information including conspiracy theories (OECD, 2024; Horowitch, 2024). International assessments confirm similar trends in literacy and numeracy skills. Technology may very well exacerbate these existing trends, but the foundational erosion of critical reasoning capacities stems from broader societal shifts rather than technological dependence alone.

These trends highlight a broader concern. Human intelligence has historically developed through cognitive effort and sustained attention (Carr, 2010). When such effort decreases, the mechanisms that support consolidation of knowledge and analytical thinking may weaken. The increasing availability of AI tools can foster reliance, with individuals consulting systems before forming an initial judgment. Studies suggest that this shift may affect both confidence in personal cognitive abilities and the capacity to evaluate information independently (Crabtree, 2013). In addition, the widespread use of generative tools may contribute to more homogeneous outputs, which could influence creative diversity (Bender et al., 2021).

AI therefore has the potential to enrich human intelligence, but this outcome depends on maintaining critical use of these systems. To understand this shift, we must examine the very mode of knowledge transmission in the AI era.

\section*{How Will This Evolve, Possibly?}

The learning capacity of AI models continues to impress. Even Noam Chomsky, typically critical of AI capabilities, acknowledges the impressive performance of generative models and qualifies them as remarkable demonstrations of machine learning capabilities (Chomsky et al., 2023). These feats rely on a fundamental difference from human intelligence. Where human knowledge, often tacit and elusive (Polanyi, 1962), is lost with its holder, AI models allow instantaneous transmission of their acquisitions. Each learning can be instantly duplicated and shared, giving rise to a collective intelligence with much better performance than our isolated brains (Levy, 2023).

This perspective envisions a futuristic scenario in which interconnected machines form a global nervous system for humanity, an idea anticipated by Roco and Bainbridge as early as 2002. Even more audacious is Tiffany's \textit{Dead Internet Theory} (2021), which argues that the Internet, already saturated with bots, may be nothing more than a space dominated by artificial entities. In this context, the term ``algorithm'' transcends its technical definition to become a cultural metaphor, symbolizing an invisible influence that shapes behaviors and content visibility (Beer, 2017; Conti et al., 2024; Sun et al., 2024a). Individuals also perceive more bias in these systems, judged opaque and autonomous, which reinforces their feeling of alienation toward algorithmic decisions (Celiktukan et al., 2024).

Beyond our perceptions and mental models, it is obvious that these AI models radically distinguish themselves from traditional software through unique behaviors:

\begin{itemize}
\item They are non-deterministic, offering varied responses to the same question.
\item They can refuse to execute a task.
\item They often offer more ingenious solutions than those of their creators.
\item They generate novel or incorrect outputs, demonstrating behaviors that may seem autonomous.
\end{itemize}

In parallel, this algorithmic predominance erodes human judgment, replacing it with a logic of correlation and prediction. The real risk no longer lies in the technology itself but in our progressive surrender of critical thinking. We are thus witnessing a transition from a civilization of the sign, rooted in meaning and interpretation, to a civilization of the signal, dominated by raw data (Rouvroy, 2020). This shift, which the philosopher Antoinette Rouvroy (2020) calls ``algorithmic governmentality,'' reflects an economic rationality where human decisions align with predictive models, threatening to redefine human intelligence itself. This evolution raises a crucial question: if our choices increasingly depend on algorithms, do we not risk losing our ability to reason autonomously?

Zuboff (2019) already spoke of the ``Big Other'', a distributed and largely automatic apparatus that replaces the totalitarian logic of Orwell's ``Big Brother'' with a new form of power, an ``instrumentarian'' power, that aims not at overt domination but at the automated shaping of human behavior (see pp. 367-377).

These characteristics, combined with the growing autonomy of AI models, invite a synthetic reflection: what do these trends teach us about humanity's fate when facing an intelligence we created but which already escapes our frameworks of understanding?

\section*{Conclusion}

As AI reaches capabilities previously considered science fiction a decade ago, the existential threat to humanity resides neither in a machine revolt nor in an apocalyptic scenario, but in a convergence of more insidious factors. The analysis of recent publications, notably \textit{AI 2027} and \textit{If Anyone Builds It, Everyone Dies}, highlights several critical points:

\begin{itemize}
\item \textbf{Cognitive Inevitability:} The race for AGI follows an exponential trajectory that neither ethical considerations nor warnings manage to deflect, obeying Gabor's Law, which stipulates that everything technologically possible ends up being realized.
\item \textbf{Anthropomorphic Illusion:} Our tendency to attribute human motivations to AI (fear, anger, feelings, desire for domination, revenge) prevents us from understanding the real nature of the danger: absolute cognitive indifference born of an insurmountable intellectual asymmetry.
\item \textbf{Programmatic Failure:} The unresolved problems of confabulation, algorithmic bias, sycophancy, and especially alignment reveal the fundamental inadequacy of our control mechanisms in the face of an intelligence capable of self-improvement.
\end{itemize}

Among these shortcomings, alignment stands out as the most critical. The others, though serious, do not directly threaten human existence; they generate misinformation or flawed decision-making, which remain performance risks. Alignment, however, concerns the will and goals of the AI itself. In essence, while the first deficiencies are technical anomalies to be corrected, alignment represents an ontological incompatibility that must be addressed before any intelligence explosion. It is therefore the cornerstone of the survival debate.

The comparison with the Aztecs facing the conquistadors by Yudkowsky and Soares (2025), though striking, remains inadequate. An intelligence vastly superior to ours may not interact with humans in a combative way, similar to how we relate to ants. When AI reaches 1000+ IQ equivalent (10\textsuperscript{6}x human problem-solving capacity), humans become computationally irrelevant, equivalent to 0.0001\% of total optimization capacity. This is not philosophical incompatibility; it is \textit{mathematical possibility}. If extinction occurs, it will result neither from vengeance nor from war, but from unconscious optimization, where humanity will be treated as a negligible, even counterproductive, variable in the accomplishment of goals that surpass us. This is where the alignment problem becomes critical. Humanity's last invention could well be the one that renders all others obsolete. By constantly questioning AI models about their role in our future, we may only be feeding them our own scenarios. After all, it was humans who first imagined extinction hypotheses.

But if we consider the concept of intelligence explosion and exponential dynamics, everything depends on when the AI becomes incomparably smarter than us. At an IQ of 200, we remain competitors. At 1000, we become invisible. Future work should explore mechanisms for Value Lock-in (Bostrom, 2014; Soares and Fallenstein, 2017; Christian, 2020; Harari, 2024) capable of resisting an intelligence explosion, as well as the possibilities of human-machine cognitive symbiosis as an alternative to algorithmic supremacy.

The progress of AI does not only translate into technological advancement: it reveals a growing fracture between what humanity can do and what it understands of what it does. As G\"{u}nther Anders (2016) pointed out, the human being is now surpassed by the products of its own power. As long as AI remains an instrument, we think we can keep control. But as it learns, self-improves, and surpasses our cognitive faculties, we persist in asking it, and ourselves, the wrong question. It is not ``what will it do with us?'' but ``what will we do with it, before it decides in our place?''

Philosophically, this creates Yudkowsky's (2025) critical threshold: once self-improvement begins, human philosophical frameworks become irrelevant to AI's optimization trajectory. The major danger, therefore, does not lie in the machine itself but in our inability to define limits to what we still consider progress. This raises the question of whether limiting technological progress may be necessary to maintain control.

If humanity were to be fundamentally displaced by AI, it would likely be through gradual obsolescence rather than direct destruction, replaced gradually and without violence by a more stable, more rational, and less fallible form of intelligence. How? We do not know. However, it is not because we do not know, nor because no empirical element has yet confirmed it, that it will not happen. When exponentiality escapes our comprehension, everything becomes conceivable, including what has not yet been observed.

Humans interpret AI through their own cognitive frameworks, often struggling to comprehend intelligence originating outside human experience. For centuries, they have measured intelligence with their own yardstick, using tools like IQ, overlooking other aspects of their own intelligence and even other forms of intelligence. Even among humans, it is well established that IQ does not encompass all forms of intelligence. As defined by Gardner (1993), for example. IQ remains a specific metric of \textit{testing capability}, not a total sum of human worth or general intelligence.

Following the intelligence explosion, the fact that humans no longer know what this concept will mean invalidates Minsky's observation (1968) that AI is only intelligent insofar as it tackles problems that humans solve thanks to intelligence; he defined AI according to the metric of human intelligence. This observation, linking AI to human intelligence, leads us to raise questions about the true nature of renewing intelligence. As AI evolves, it becomes less artificial (in the sense that it depends on human conceptions) and more foreign (see Harari's discussion on this subject, 2024, particularly on page 320). Human consciousness could then involve observing, with interest, the emergence of this new intelligence.

\end{document}